\documentclass{article} %
\usepackage{iclr2025_conference,times}
\usepackage{graphicx}

\usepackage{amsmath,amsfonts,bm}

\def\eqref#1{equation~\ref{#1}}

\def\1{\bm{1}}

\DeclareMathAlphabet{\mathsfit}{\encodingdefault}{\sfdefault}{m}{sl}
\SetMathAlphabet{\mathsfit}{bold}{\encodingdefault}{\sfdefault}{bx}{n}

\usepackage{hyperref}
\usepackage{url}
\usepackage{algorithm}
\usepackage{algpseudocode}
\usepackage{ragged2e}

\title{Atlantes: A system of GPS transformers for global-scale real-time maritime intelligence}

\author{Henry Herzog, Joshua Hansen, Yawen Zhang, \& Patrick Beukema  \\
Allen Institute for AI (Ai2) \\
Seattle, WA, USA \\
\texttt{\{henryh,joshuah,yawenz,patrickb\}@allenai.org} \\
}

\iclrfinalcopy %
\begin{document}

\maketitle

\begin{abstract}
Unsustainable exploitation of the oceans exacerbated by global warming is threatening coastal communities worldwide. Accurate and timely monitoring of maritime activity is an essential step to effective governance and to inform future policy. In support of this complex global-scale effort, we built \textit{Atlantes \footnote{Our base model architecture is called ATLAS (AIS transformers learning for active subpaths). \textit{Atlantes} refers to a collection of atlases.}}, a deep learning based system that provides the first-ever real-time view of vessel behavior at global scale. Atlantes leverages a series of bespoke transformers to distill a high volume, continuous stream of GPS messages emitted by hundreds of thousands of vessels into easily quantifiable behaviors. The combination of low latency and high performance enables operationally relevant decision-making and successful interventions on the high seas where illegal and exploitative activity is too common. Atlantes is already in use by hundreds of organizations worldwide. Here we provide an overview of the model and infrastructure that enables this system to function efficiently and cost-effectively at global-scale and in real-time. 
\end{abstract}
\section{Introduction}
\begin{figure}[t]
  \centering
   \includegraphics[width=1.0\linewidth]{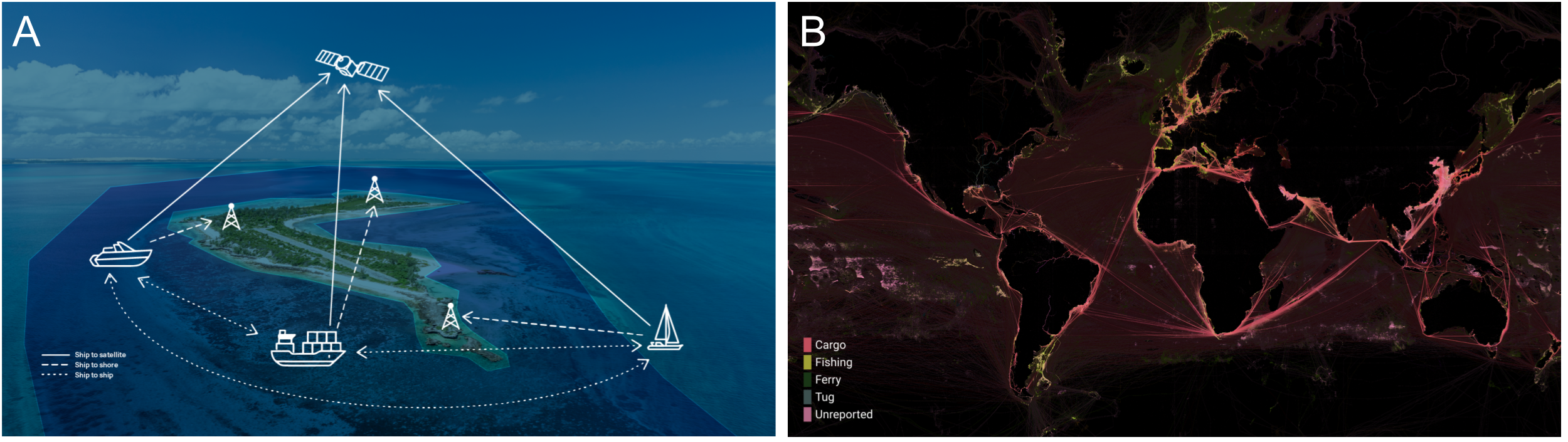}
   \caption{(A) Depiction of the Automatic Identification System. (B) All broadcasted messages from 2023 ($\sim$ 15B) color coded by vessel category.}
   \label{fig:overview}
   \vspace{-4mm}
\end{figure}

Each day, approximately 600,000 vessels use the Automatic Identification System (AIS) to transmit a stream of GPS messages—each containing a unique identifier, timestamp, latitude and longitude, and often metadata such as speed and heading—to a constellation of satellites. These GPS streams are delivered to a diverse set of stakeholders, including coastal-state governments, environmental organizations, and commodities traders (Fig. \ref{fig:overview}A). Although AIS was originally conceived to enhance collision avoidance, it now underpins a wide range of applications: supply-chain tracking, route optimization, insurance and compliance, commodity monitoring, environmental surveillance, and more. Alongside satellite imagery, AIS constitutes one of the two principal datasets for global maritime intelligence—yet unlike imagery’s intermittent snapshots, AIS provides a near-comprehensive, real-time feed of vessel movements worldwide.

In this paper, we describe a new suite of models that makes this information useful in operational contexts for real-time decision making. Our primary goal was to build an application \citep{rolnick2024} that could meet users' needs -- minimal latency and expert performance. Additionally, the model had to be highly computationally efficient to be cost-effective to run in production and to enable high velocity iteration. To achieve this, we trained several transformer-based architectures via supervision on a large dataset of GPS trajectories annotated at message-level granularity by a group of leading maritime analysts. The entire code base has been open-sourced, including training and inference code, as well as the service production code, allowing users to deploy this model or fine-tune it for other use cases (\href{https://github.com/allenai/atlantes}{GitHub}).

Prior research on AIS data has primarily concentrated on specific geographic regions or short temporal windows, with an emphasis on retrospective analysis rather than real-time classification (see \cite{YANG2024103426} and \cite{jmse10010112} for comprehensive reviews). In 2018, Global Fishing Watch introduced a convolutional neural network that generates historical classifications of global vessel traffic using twelve handcrafted features, including variations in speed and course \citep{kroodsma2018tracking}. \citet{liang2022} proposed a scalable, transformer-based architecture for classifying urban transportation modes, which inspired our approach. The lack of mature, GPS-specific machine learning toolkits—whether commercial or open source—has posed a challenge to the development of scalable and robust models. Beyond AIS data, research into modeling GPS trajectories for human mobility is gaining interest, with emphasis on the need for new foundation models \citep{choudhury24}. 

\section{Methods}
\subsection{Tasks and Datasets}
Atlantes tackles two classification tasks with AIS sequences: (1) entity classification to determine whether a sequence is from a vessel or a buoy, and (2) activity classification to determine the vessel's activity given the most recently broadcasted AIS message. Entity classification is a requirement because without that context, it is not possible to distinguish activity patterns on the basis of the most recently broadcasted message. For example, a vessel may transport a broadcasting buoy resulting in vessel-like trajectories for a period of time, or a vessel may drift without power resembling a buoy.

\textit{Entity classification dataset}: We applied domain-expert heuristics and leveraged AIS metadata to sample buoys with high precision but low recall across the entire population of AIS from 2024 ($\sim$20B messages). The result was a dataset of 1.8M entities (175,857 buoy labels and 1,643,737 vessel labels). All buoy data was included, while vessel data was stratified by vessel type.

\textit{Activity classification dataset}: We could not rely on metadata or heuristics to annotate vessel trajectories by activity class because no existing metadata reliably separates these behaviors (even if one sacrifices significant recall). To build an effective global model we compiled a training dataset that covered the entire planet and all vessel types, continuously between January 2022 and June 2024. We hired 20 expert maritime analysts capable of distinguishing other behaviors from fishing. The analysts were trained to ensure consistent labeling against domain definition, and label every GPS message into five classes: transiting, anchored, fishing, moored, or other. This work was conducted using a custom-built platform designed specifically for GPS trajectory annotation (Fig. \ref{fig:annotation}). During the initial sampling phase, 85\% of monthly stratified samples came from expert-identified regions, with the remaining 15\% from other areas. These data were used to train an initial model, which was later improved via active learning. Experts iteratively labeled new training samples informed by the model predictions. In total, experts labeled over 7,500 track-months ($\sim 10k$ messages in a track-month) of AIS data, encompassing more than 15 million messages.  These labels include 558,347 instances of fishing, 502,490 of transiting, 123,696 of moored, and 105,299 of anchored activities. 

\begin{figure}[h]
  \centering
   \includegraphics[width=1.0\linewidth]{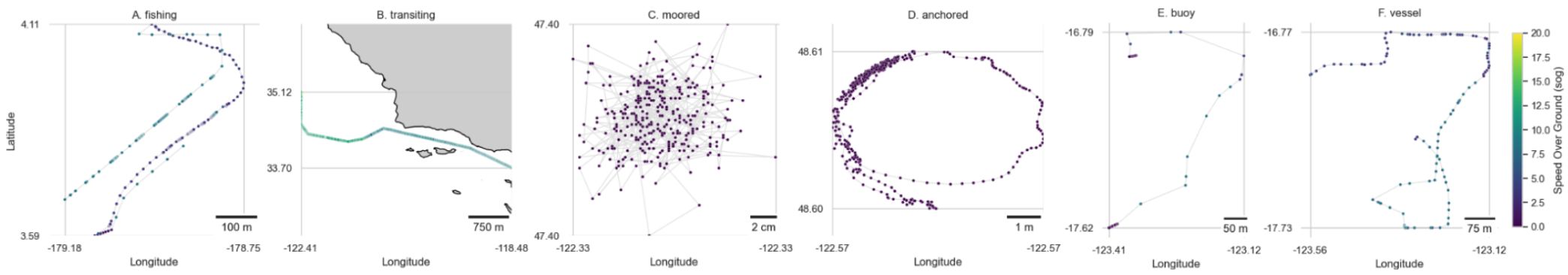}
   \caption{Examples of activity from each class (A-D). Example buoy (E) and vessel (F).} 
   \label{fig2}
   \vspace{-4mm}
\end{figure}

\subsection{Model architecture and training}
We found that transformers are well suited to AIS modeling, but several critical differences to natural language motivated a bespoke approach. Unlike natural language, AIS/GPS sequences exhibit highly irregular spatial and temporal frequencies, and can contain high amounts of noise. This irregularity is handled by a continuous point embedding layer (CPE) as described in \cite{liang2022}, coupled to CNN layers, and the transformer encoder. The CPE layers embed the raw GPS sequences, which are irregular in space and time, into a representation suitable for the encoder layers. This is achieved by computing the spatiotemporal differences between successive anchor points within a configurable window (n=9 messages). This step eliminates the need for message interpolation or extensive feature engineering. Subsequent CNN layers extract local patterns, and the transformer encoder learns global representations across the entire sequence via multi-head self-attention. Two separate models were trained using the ATLAS architecture, one for entity classification and a second for activity classification. The ATLAS architecture includes 6 CPE layers, 3 CNN layers, and 9 transformer layers for activity classification, 4 for entity classification. In addition to the model, activity classification results are passed through several post-processing steps, including confidence thresholding, geo-fencing around marine infrastructure \citep{bastani2023}, and speed filters. 

Models were trained using class-weighted cross-entropy loss, with weights selected based on the minimum validation loss over 4 epochs, using a batch size of 256, and a learning rate of 1e-4. A single training run takes just 6 hours on four H100 GPUs. The lightweight design of the model, consisting of 4.7M parameters, makes it computationally efficient and suitable for deployment on modest hardware, and self hosting.  

\begin{figure}[h]
  \centering
   \includegraphics[width=1.0\linewidth]{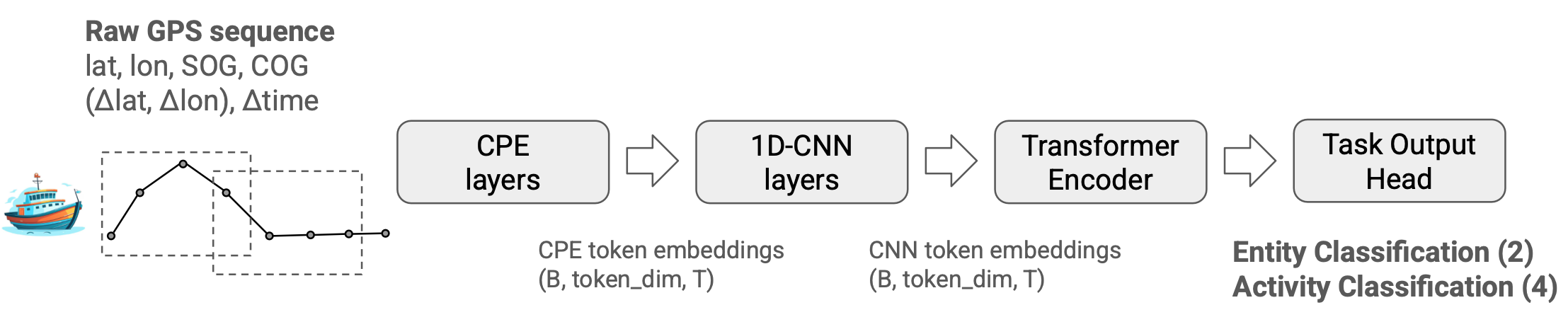}
   \caption{\textbf{\texttt{ATLAS}} model architecture (SOG: speed over ground, COG: course over ground).} 
   \label{fig:model}
   \vspace{-4mm}
\end{figure}

\subsection{How good is this model really?}
While we conducted extensive offline evaluations, our primary evaluation method involved experts reviewing the model's outputs in an online staging environment (prior to exposing results to all users). After addressing different sources of false positives identified through this feedback process, we conducted third-party evaluation on globally stratified random samples. For entity classification, our model achieved an accuracy of 97.5\%. For activity classification, our model achieved an overall accuracy of 71\% (90\% for fishing vessels and 51\% for unknown vessels). Because there is no other model that offers real-time global predictions, we are not able to produce a direct comparison of this model to prior work. We share these metrics because they best reflect existing user experience. 

The best context we can provide about how ``good" our performance is, is that we are approaching human performance but not quite there. We measured inter-annotator agreement on the activity classification task at 85\%. There may be multiple plausible activities that will only converge after a more prolonged period of engaging in that activity. Humans also typically use many different data sources, in addition to the GPS sequence, to classify a trajectory. The model only has access to the GPS stream. Naturally, performance improves with additional context. At the extreme, classification of GPS messages conditioned on future data, such as analysis of historical data, is a different problem, and should approach perfect performance.

\subsection{Deployment}
Upon receiving new messages, a speed-based change-point detection algorithm identifies any putative change in behavior, triggering a three-stage pipeline involving pre-processing, the model inference, and post-processing (see Appendix). Entity classification is performed offline using historical AIS data stored in a database and is only calculated if metadata for the activity classification tasks lacks that class. This information is typically missing for new trajectories or vessels not previously on the water broadcasting. If insufficient context exists (n=500 messages), the entity classifier defaults to unknown. For activity classification, entity information determines whether specific activities are possible (e.g., fishing events are only associated with vessels). 

The system processes 28 activity classifications per second using 20 CPUs and 5 T4 GPUs, with a configurable message context of up to 2048 messages spanning 30 days (approximately 5B GPS messages are inferenced including the context each day).

\begin{figure}[h]
  \centering
   \includegraphics[width=.7\linewidth]{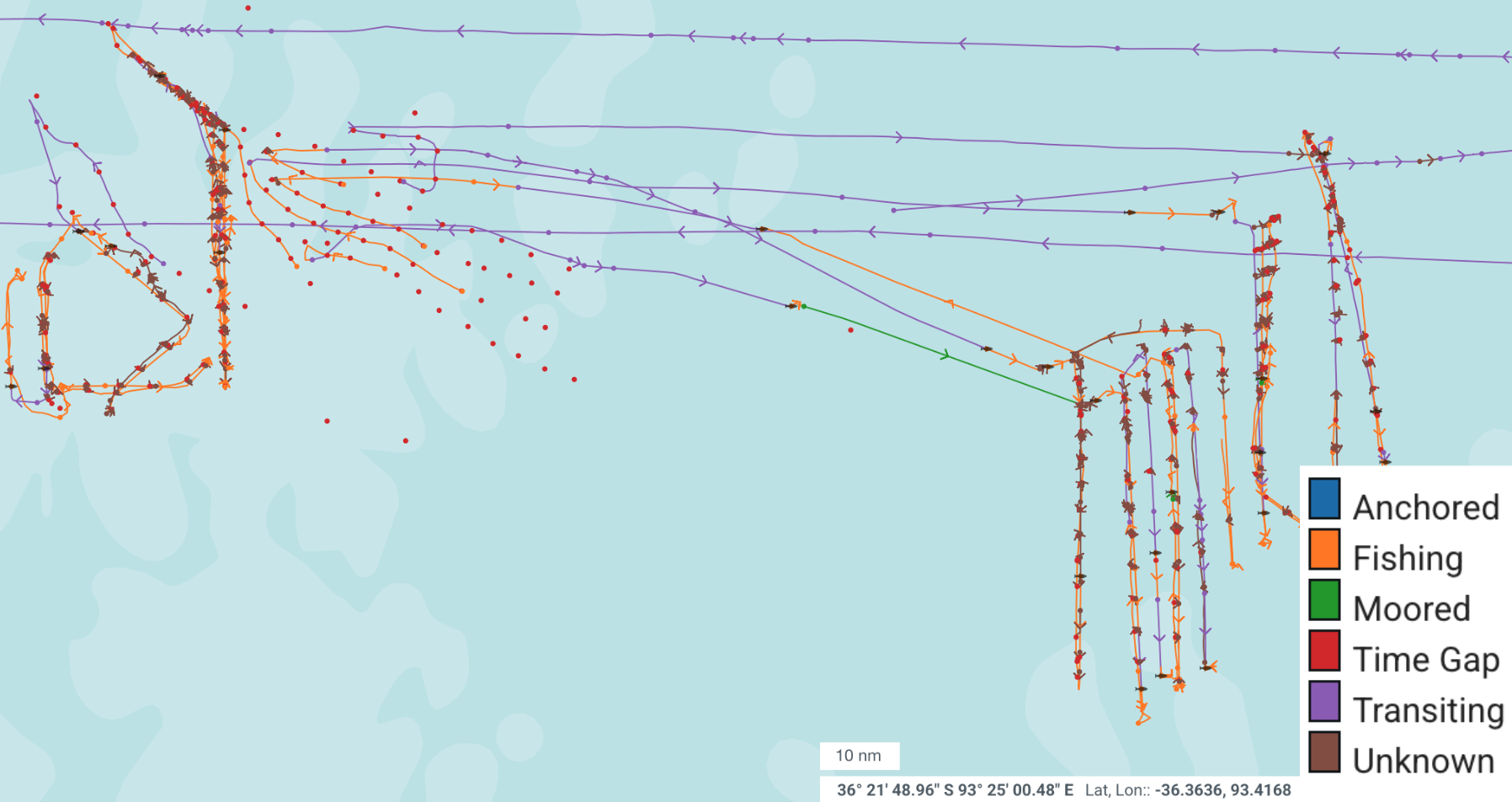}
   \caption{Example classifications of vessel movement patterns (Indian ocean, February 2025).} 
   \label{fig:example_output}
   \vspace{-5mm}
\end{figure}

\section{Conclusion/Impact Statement}

Our primary contribution is to enable real-time analysis and therefore actionable intelligence. Real-time insights require classifying behavior with limited information, and care must be taken to ensure that the intelligence we provide is both accurate and worth taking action on. This model is merely one component in a much more complex and global effort to monitor and protect the planet's oceans. We hope that by open sourcing the models, we can enable other researchers and activists to better understand the strengths and limitations of this approach, and also foster more widespread adoption of maritime transparency.

\newpage
\subsubsection*{Acknowledgments}
We wish to thank the annotation team for curating and refining the dataset that made all of these models possible. We also thank the engineering team for their work in developing the annotation tool which enabled scalable annotations. Additionally, we thank Spire Maritime for data. Thanks to Karen Goodfellow for help with UI and UX design. 

\setlength{\bibsep}{0pt}
\bibliography{iclr2025_conference}

\begin{thebibliography}{7}
\providecommand{\natexlab}[1]{#1}
\providecommand{\url}[1]{\texttt{#1}}
\expandafter\ifx\csname urlstyle\endcsname\relax
  \providecommand{\doi}[1]{doi: #1}\else
  \providecommand{\doi}{doi: \begingroup \urlstyle{rm}\Url}\fi

\bibitem[Bastani et~al.(2023)Bastani, Wolters, Gupta, Ferdinando, and Kembhavi]{bastani2023}
Favyen Bastani, Piper Wolters, Ritwik Gupta, Joe Ferdinando, and Aniruddha Kembhavi.
\newblock Satlaspretrain: A large-scale dataset for remote sensing image understanding, 2023.
\newblock URL \url{https://arxiv.org/abs/2211.15660}.

\bibitem[Choudhury et~al.(2024)Choudhury, Kreidieh, Kuznetsov, and Arora]{choudhury24}
Shushman Choudhury, Abdul~Rahman Kreidieh, Ivan Kuznetsov, and Neha Arora.
\newblock Towards a trajectory-powered foundation model of mobility.
\newblock In \emph{Proceedings of the 3rd ACM SIGSPATIAL International Workshop on Spatial Big Data and AI for Industrial Applications}, GeoIndustry '24, pp.\  1–4, New York, NY, USA, 2024. Association for Computing Machinery.
\newblock ISBN 9798400711459.
\newblock \doi{10.1145/3681766.3699610}.
\newblock URL \url{https://doi.org/10.1145/3681766.3699610}.

\bibitem[Kroodsma et~al.(2018)Kroodsma, Mayorga, Hochberg, Miller, Boerder, Ferretti, Wilson, Bergman, White, Block, et~al.]{kroodsma2018tracking}
David~A Kroodsma, Juan Mayorga, Timothy Hochberg, Nathan~A Miller, Kristina Boerder, Francesco Ferretti, Alex Wilson, Bjorn Bergman, Timothy~D White, Barbara~A Block, et~al.
\newblock Tracking the global footprint of fisheries.
\newblock \emph{Science}, 359\penalty0 (6378):\penalty0 904--908, 2018.

\bibitem[Liang et~al.(2022)Liang, Ouyang, Wang, Liu, Chen, Zhang, Zheng, and Zimmermann]{liang2022}
Yuxuan Liang, Kun Ouyang, Yiwei Wang, Xu~Liu, Hongyang Chen, Junbo Zhang, Yu~Zheng, and Roger Zimmermann.
\newblock Trajformer: Efficient trajectory classification with transformers.
\newblock In \emph{Proceedings of the 31st ACM International Conference on Information \& Knowledge Management}, CIKM '22, pp.\  1229–1237, New York, NY, USA, 2022. Association for Computing Machinery.
\newblock ISBN 9781450392365.
\newblock \doi{10.1145/3511808.3557481}.
\newblock URL \url{https://doi.org/10.1145/3511808.3557481}.

\bibitem[Rolnick et~al.(2024)Rolnick, Aspuru-Guzik, Beery, Dilkina, Donti, Ghassemi, Kerner, Monteleoni, Rolf, Tambe, and White]{rolnick2024}
David Rolnick, Alan Aspuru-Guzik, Sara Beery, Bistra Dilkina, Priya~L. Donti, Marzyeh Ghassemi, Hannah Kerner, Claire Monteleoni, Esther Rolf, Milind Tambe, and Adam White.
\newblock Application-driven innovation in machine learning, 2024.
\newblock URL \url{https://arxiv.org/abs/2403.17381}.

\bibitem[Wolsing et~al.(2022)Wolsing, Roepert, Bauer, and Wehrle]{jmse10010112}
Konrad Wolsing, Linus Roepert, Jan Bauer, and Klaus Wehrle.
\newblock Anomaly detection in maritime ais tracks: A review of recent approaches.
\newblock \emph{Journal of Marine Science and Engineering}, 10\penalty0 (1), 2022.
\newblock ISSN 2077-1312.
\newblock \doi{10.3390/jmse10010112}.
\newblock URL \url{https://www.mdpi.com/2077-1312/10/1/112}.

\bibitem[Yang et~al.(2024)Yang, Liu, Li, Zhang, and Liu]{YANG2024103426}
Ying Yang, Yang Liu, Guorong Li, Zekun Zhang, and Yanbin Liu.
\newblock Harnessing the power of machine learning for ais data-driven maritime research: A comprehensive review.
\newblock \emph{Transportation Research Part E: Logistics and Transportation Review}, 183:\penalty0 103426, 2024.
\newblock ISSN 1366-5545.
\newblock \doi{https://doi.org/10.1016/j.tre.2024.103426}.
\newblock URL \url{https://www.sciencedirect.com/science/article/pii/S1366554524000164}.

\end{thebibliography}
\bibliographystyle{iclr2025_conference}

\appendix
\section{Appendix}
\subsection{Change point detector logic summary}
The change point detector identifies significant shifts in observational data based on time and speed over ground (SOG) metrics. It operates under the assumption that the last element in the input list represents the most recent observation. By combining time-based and SOG-based analyses, the detector determines if a changepoint has occurred and clearly communicates the reasoning behind each detection outcome. See GitHub for a complete description of the algorithm: \url{https://github.com/allenai/atlantes/blob/main/ais/src/atlantes/cpd/changepoint_detector.py}

\subsection{Annotation Tool}

While developing Atlantes, we developed a custom machine learning annotation tool designed to streamline and scale GPS annotations. Conventional GPS annotation methods relied on ad-hoc tooling and did not scale robustly. Our in-house tool addresses these challenges by providing an intuitive, efficient platform for subject matter experts to visualize, evaluate, and label AIS data.



\begin{figure}[h]
  \centering
   \includegraphics[width=.5\linewidth]{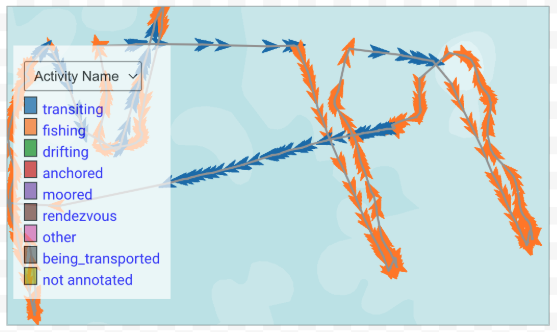}
   \caption{A. Depiction of annotation process. GPS sequences are annotated at message level granularity by assigning a class to each individual message.  } 
   \label{fig:annotation}
\end{figure}

\begin{figure}[h]
  \centering
   \includegraphics[width=.75\linewidth]{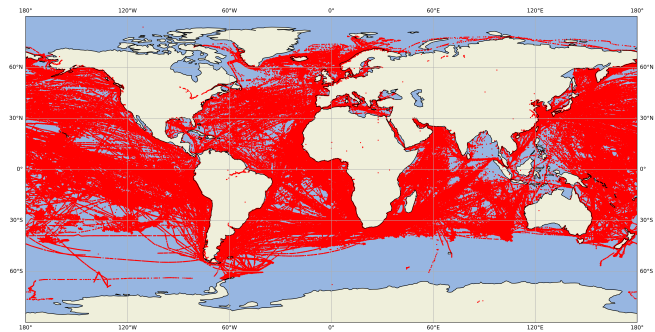}
   \caption{Geographic distribution of labeled messages used for activity classification.} 
   \label{fig:onecol}
\end{figure}

\subsection{Additional Examples of Activity Predictions}

\begin{figure}
    \centering
    \includegraphics[width=1\linewidth]{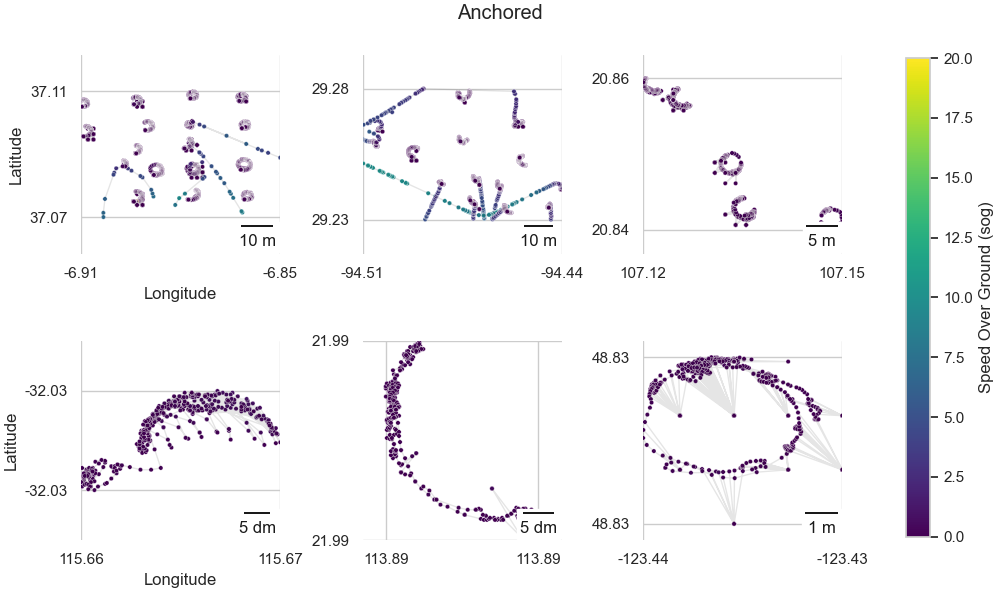}
\end{figure}

\begin{figure}
    \centering
    \includegraphics[width=1\linewidth]{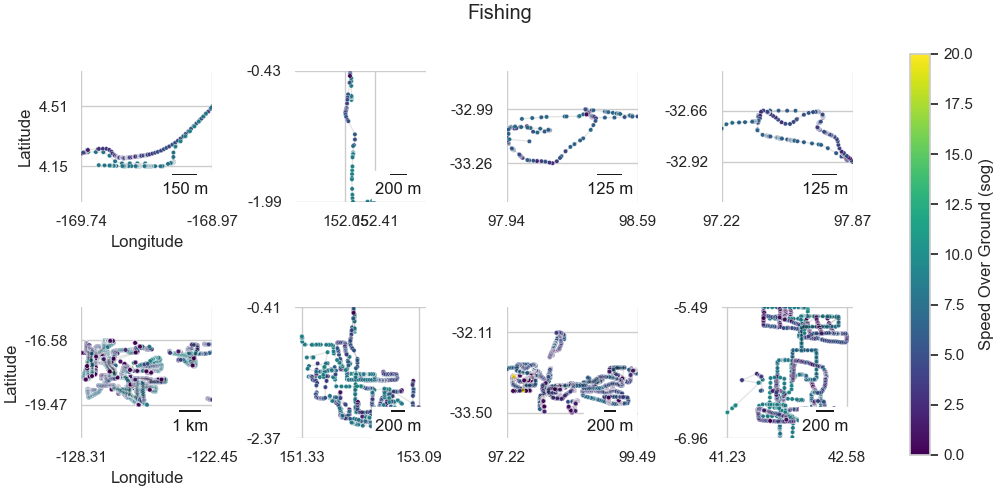}
\end{figure}

\begin{figure}
    \centering
    \includegraphics[width=1\linewidth]{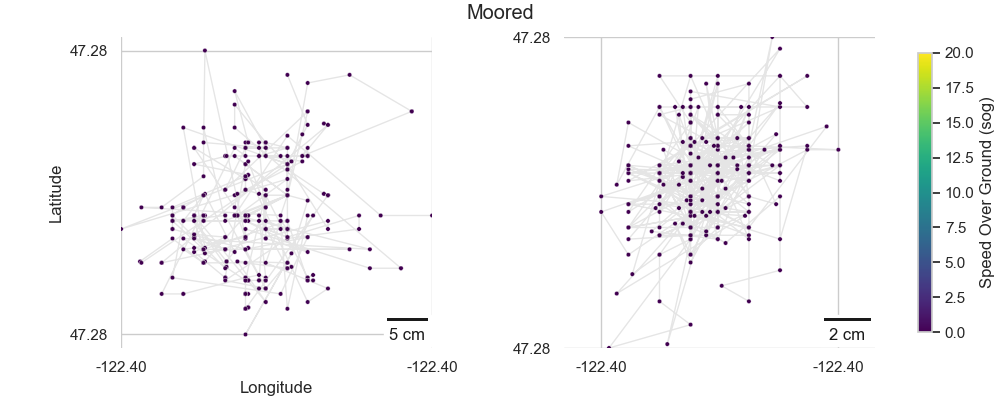}
\end{figure}

\begin{figure}
    \centering
    \includegraphics[width=1\linewidth]{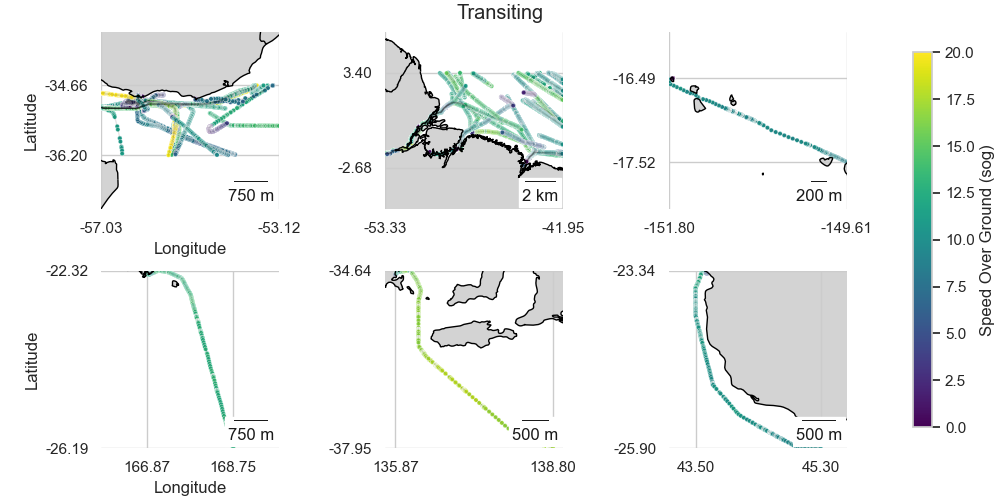}
\end{figure}

\begin{table}
    \centering
    \begin{tabular}{cc}
         Anchored& 2.26/sec\\
         Fishing& 0.47/sec\\
         Moored& 12.4/sec\\
         Transiting& 15.5/sec\\
         Unknown& 8.5/sec\\
 Total&39.1/sec\\
    \end{tabular}
    \caption{Inference rates by activity type.}
    \label{tab:my_label}
\end{table}

\end{document}